\documentclass[10pt,twocolumn,letterpaper]{article}

\usepackage{wacv}
\usepackage{times}
\usepackage{epsfig}
\usepackage{graphicx}
\usepackage{amsmath}
\usepackage{amssymb}
\usepackage{booktabs}

\wacvfinalcopy 

\def\wacvPaperID{656} 

\ifwacvfinal\fi
\begin{document}

\title{Learning to Localize and Align Fine-Grained Actions to Sparse Instructions}

\author{Meera Hahn \\
Georgia Institute of Technology\\
{\tt\small meerahahn@gatech.edu}
\and
Nataniel Ruiz \\
Boston University\\
{\tt\small nruiz9@bu.edu}
\and
Jean-Baptiste Alayrac \\
PSL Research University \\
Inria \\
{\tt\small jean-baptiste.alayrac@inria.fr}
\and
Ivan Laptev \\
Inria \\
{\tt\small ivan.laptev@inria.fr}
\and
James M. Rehg \\
Georgia Institute of Technology\\
{\tt\small rehg@gatech.edu}
}

\maketitle
\ifwacvfinal\thispagestyle{empty}\fi

\begin{abstract}
Automatically generating textual video descriptions that are time-aligned with the video content is a long-standing goal in computer vision. The task is challenging due to the difficulty of bridging the semantic gap between the visual and natural language domains. This paper addresses the task of automatically generating an alignment between a set of instructions and a first person video demonstrating an activity. The sparse descriptions and ambiguity of written instructions create significant alignment challenges. The key to our approach is the use of egocentric cues to generate a concise set of action proposals, which are then matched to recipe steps using object recognition and computational linguistic techniques. We obtain promising results on both the Extended GTEA Gaze+ dataset and the Bristol Egocentric Object Interactions Dataset.
\end{abstract}

\section{Introduction}
\label{sec:intro}
Humans and robots rely on experts to teach them how to perform complex tasks and activities. Teaching through hands-on demonstration is a classical and effective approach, but it is fundamentally not scalable. The advent of video sharing sites such as YouTube has created a boom in the development of instructional videos, which allow experts to share recorded demonstrations with a global audience. Instructional videos are also an attractive resource for autonomous learning by robots and intelligent agents, because they directly showcase the patterns of movement and tool use which define complex actions, and they connect to a long history of research in programming by demonstration~\cite{argall2009survey} and imitation learning~\cite{Schaal1999}. When instructional videos are not available or not accessible, people often use recipes to help guide them through performing unfamiliar activities. A recipe for a specific task contains much sparser information than videos but for this reason are able to condense key information about a task and generalize over many demonstration videos.

While textual instructions are compact and easily searchable they are also naturally sparse and therefore require inference of unexplict steps. In contrast, videos contain a wealth of fine-grained detail, but suffer from an inherent linearity that makes it difficult to extract the ``big picture'' view of an activity or quickly find a specific step. Since instruction sets, which we will henceforth refer to as recipes, and instructional videos are complementary sources of information, it is beneficial to create a system that can automatically align a video and a corresponding recipe. 

One way to bridge instructional videos and text is through narration~\cite{alayrac2016unsupervised,bojanowski2015weakly}, which when transcribed brings a sequence of words into alignment with the video frames. However the scalability of this approach is limited by the relative scarcity of text transcribed narrations of instructional videos. Therefore we focus on non-narrated videos and recipes. We also choose to work with First Person (FP) videos, which are particularly attractive for activity understanding, as a headworn camera is ideally positioned to capture the actions of the camera wearer, and FP videos are becoming increasingly common in social media~\cite{Nguyen2016}. Recent work has demonstrated the utility of the first person approach in activity recognition~\cite{Ma_2016_CVPR,li2015delving}.

\begin{figure*}
\label{teaser}
  \centering
    \includegraphics[scale=.23]{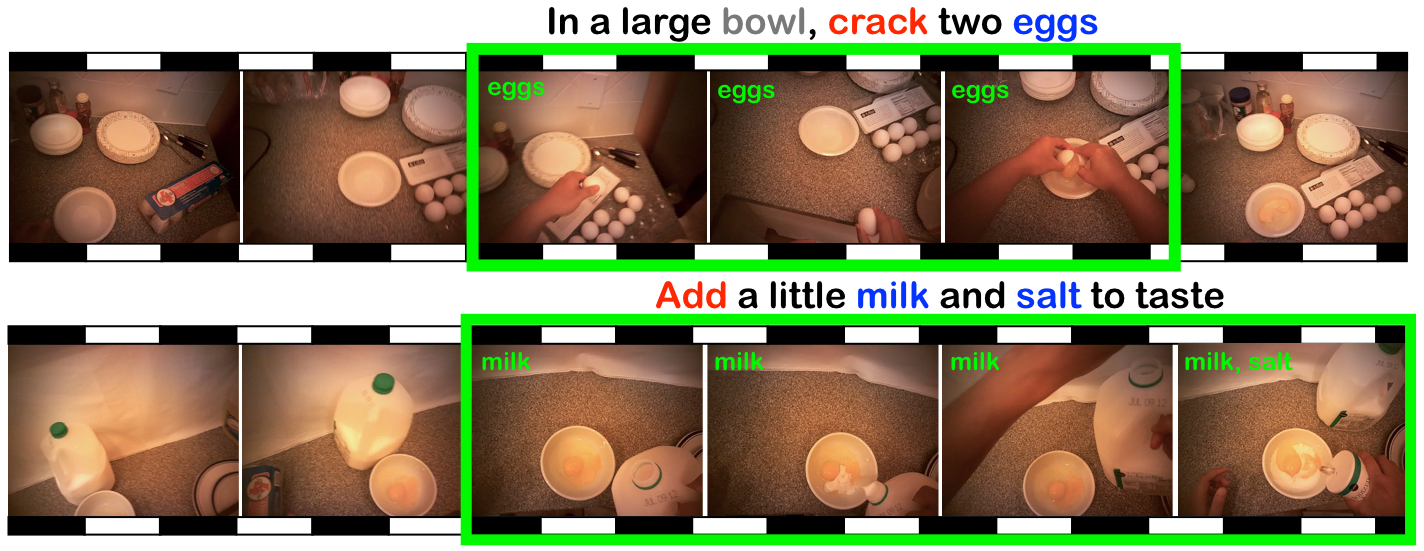}
  \caption{The final results of the system on EGTEA dataset. Video frames that are surrounded by a green box are part of a predicted action segment. The green words in the corner of the frames in the action segments are the recognized objects in the frame. The sentence above the action segment is the recipe step that the system automatically aligned the segment to. Each recipe step is colored according to how it was parsed by our alignment module. Blue words are primary objects, gray words are secondary objects and red words are actions.}
\end{figure*}

In this paper, we propose a novel task of recipe to egocentric video alignment, in which a \emph{non-narrated} first person video recording of an activity is \emph{aligned} with the steps in the recipe that describes it. This means that for every instructional step in a recipe all clips in the input video that correspond to that step are automatically temporally localized and aligned to that step. Additionally no two recipe steps can be aligned to the same video segment. An example of the task illustrated in Figure \ref{teaser}. The two key technical challenges of this alignment task are 1) the extremely low ratio of text to video; and 2) the fact that no explicit correspondence is given at training or testing time between the recipe text and the video frames, making the alignment module unsupervised. In contrast, most previous tasks concerning video annotation or instructional video analysis either assume that frame-level text annotations are available \cite{xu2015jointly,densecaptioning,tacos} or infer them from dense text descriptions obtained from transcribing video narrations \cite{alayrac2016unsupervised,bojanowski2015weakly}. The nature of recipe text to be sparse is a key factor underlying the difficulty of the alignment task that we propose. 

To address the challenges of the task we adopt a two-stage approach in which we first extract all the segments from the video that have a high probability of containing an action and then align the resulting action proposals with the steps in the recipe. In the first stage of our approach generate action proposals over the video specifically by exploiting the \emph{egocentric cues} of first person videos, such as hand and head movement along with eye gaze. The action proposals identify video segments where objects are manipulated by some action. During the second stage of our approach we automatically detect these manipulated objects in order to generate recipe alignments. The detected objects provide a link between a video clip and the steps in the recipe. The sparseness of the recipe's text descriptions leads to significant ambiguity in segment to step matching, which we address through the use of unsupervised linguistic reference resolution. We test the resulting system for aligning fine-grained actions to recipe steps by evaluating it on an multiple FP video datasets. We obtain promising results, outperforming both recent action proposal methods and prior video-text alignment pipelines. Our approach has the potential to unlock the vast repository of textual descriptions of actions by aligning them with the growing corpus of FP videos, enabling new approaches for human instruction and robot learning. Our system can also be used for video retrieval making instructional videos more accessible.


This paper makes three contributions:  First, we present a system for automatically generating action proposals in FP videos which exploits egocentric cues for localizing the onset and offset of actions. Second, we create a novel method for aligning action proposals to steps in a recipe via action and object-derived features. Third, we annotate the popular EGTEA Gaze+ dataset and the Bristol Egocentric Object Interaction Dataset for the new recipe to ego-video alignment task and we show experimental results over them.

\section{Related Works}
\label{sec:relatedworks}
\noindent \textbf{Text-to-Video:}
The key difference between this paper and past work in text-to-video alignment ~\cite{alayrac2016unsupervised,xu2015jointly,bojanowski2015weakly,malmaud2015s}, is that these works rely on dense text that is curated for each individual video. In contrast, the task we have outlined is to align multiple videos to the single recipe that describes the activity, in this paper we present a system that is able to accomplish this task.  One of the previous methods~\cite{alayrac2016unsupervised} identifies the key steps in an instructional video by clustering dense, transcribed narrations, and then aligns each step to a video in a weakly-supervised end-to-end method. \cite{alayrac2016unsupervised} proves that narrations are an informative signal, however narrations are not always available and spoken narrations are not always accurately transcribed by automatic speech recognition systems. \cite{huang2017unsupervised} aims to resolve the noise in narrations using a visual-linguistic model but does not solve the alignment task. In fact the system in \cite{huang2017unsupervised} relies on the narrations already being aligned to the frames. \cite{bojanowski2015weakly} performs weakly-supervised alignment of videos to text using the TACoS~\cite{tacos} dataset, which has captions for each video. In contrast, in the datasets used in our work there is only a single recipe for all videos of that class rather than individual recipes for each video. This means that some videos may skip recipe steps or add activities that don't correspond to any recipe step. The problem of aligning videos to recipes is first addressed in~\cite{malmaud2015s}, however they use written narrations to do the alignment. Using purely linguistic model~\cite{malmaud2015s} aligns the narrations to the recipe steps and uses visual cues to refine the alignment. In contrast, we match videos to recipes without relying on additional narrations.

\begin{figure*}[ht!]
  \centering
    \includegraphics[scale=.5]{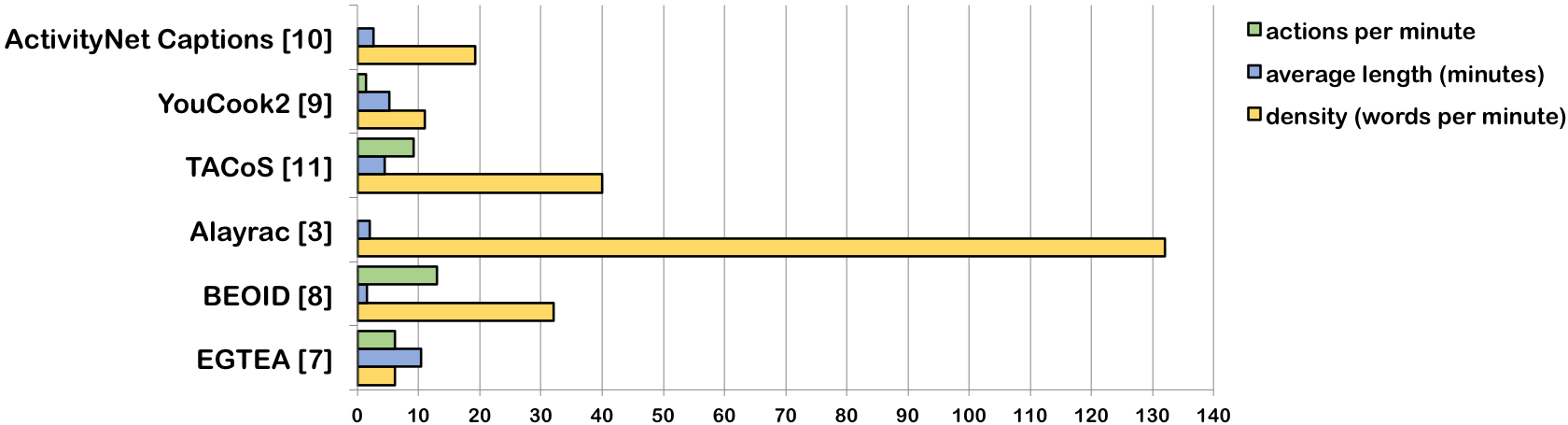}
    \label{end}
  \caption{Statistics of relevant instructional video and video captioning datasets.}
  \label{stats}
\end{figure*}

\noindent \textbf{Comparison of Sparseness in Alignment Tasks:}
In order to compare and quantify the \emph{density} of text annotations in our alignment task and previous work, we measure the average number of words available per minute of video for different datasets. We do this because text density is a key challenge of the recipe to egocentric video alignment task, with high-density datasets containing more useful information for fine-grained alignment. We evaluate our system over two egocentric datasets: Extended GTEA Gaze+ corpus (EGTEA Gaze+)~\cite{li2015delving} and the Bristol Egocentric Object Interactions Dataset~\cite{beoid}. We directly compare the density of these datasets against other instructional video and captioning datasets \cite{alayrac2016unsupervised,corso,densecaptioning,tacos} in the graph in Figure~\ref{stats}. Since recipes tend to be written in short concise sentences, EGTEA has the lowest word density at 6.18 words/minute and is followed by the other recipe dataset YouCook2 with a word density of 11 words/minute. YouCook2 however does not contain First Person videos. Another important property of our approach is that it is designed to work well on continuous, unsegmented, long videos. As we can see in the graph in Figure~\ref{stats}, video clips in EGTEA are twice as long as the clips in other datasets. 

\noindent \textbf{Action Detection and Proposal Generation:}
One of the main contributions of this paper is the action segmentation module of our system. Action proposal is a well known task in the computer vision field and is often used as part of action detection methods~\cite{yeung2016end,bilen2016dynamic,singh2016multi}. Our novel method for proposal generation in FP videos is most closely related to Deep Action Proposals (DAPs)~\cite{escorcia2016daps}. DAPs produces action proposals by sliding a window of a given step size over video, inputting C3D features of the frames in the window into a unidirectional LSTM. Instead we treat the task as frame level classification. DAPs outputs multiple overlapping candidate windows whereas our method does not allow overlapping action segments. Additionally, DAPs works best on long videos and isn't as effective at generating proposals for short videos, see \S\ref{sec:align_res} and Table~\ref{fullresults}. We differ from all previous action proposal systems by incorporating egocentric cues into our network and testing with video-to-text alignment.
The works \cite{richard2017weakly,huang2016connectionist} do weakly supervised action labeling, which takes a temporally ordered set of action labels that occur in a video and align the video's frames to the actions. Our task is different since we take in written recipes which have to be parsed and which do not explicitly state all actions that occur in a video. Also the recipes are not treated as a strictly ordered set since recipe steps can be done out of order.

\noindent \textbf{First Person Vision:}
Our system is created for first person (FP) videos which have become more prevalent in the computer vision community in recent years ~\cite{Ma_2016_CVPR,li2015delving,pirsiavash2012detecting,kitani2011fast,zhang2017first}. We utilize the egocentric cues proposed by~\cite{li2015delving} in our method for action proposal generation.~\cite{li2015delving} focuses on action recognition over predefined ground truth action segments, in contrast, we automatically generate localized segments containing actions and then match them to recipe steps. Our use of gaze-related features connects to prior work on gaze prediction in FP video~\cite{li2013learning,yamada2012attention}.

\section{Approach}

\label{sec:approach}
Given a recipe in natural language text and a corresponding egocentric video, the goal of our system is to \emph{align} the recipe and video by matching each recipe step to all of the video frames that correspond to its execution, as shown in Figure~\ref{teaser}. Our approach has three main parts: 1) Identifying video segments that contain actions and are therefore candidates for matching; 2) Parsing the recipe text into a representation of objects and actions which supports matching; and 3) Establishing the correspondence between identified segments and recipe steps via the objects that are being used and manipulated. A diagram of the complete pipeline is shown in the Figure~\ref{full}. 

The main challenge of the alignment task is that recipe descriptions only provide sparse high-level descriptions of complex activities, e.g. ``whisk a few eggs in a large bowl''. This type of abbreviated description can map onto a large number of potential sequences of atomic actions. Instead of making a detector for each possible action, we explore the hypothesis that we can identify and align relevant video segments to their recipe steps with only \emph{action proposals} and task-relevant object recognition. Our approach leverages the discriminative power of egocentric cues, such as hand configurations and head motion. We show for the first time their usefulness in temporally localizing actions. After generating action proposals for a video, we then automatically identify the main objects that are being manipulated during the actions. We use these detected objects as concrete reference points with which to establish a correspondence to specific recipe steps. The final output of the entire pipeline is a set of video segments for each step in the recipe, where the segments delineate the actions described in that particular recipe step.

\begin{figure*}
  \centering
    \includegraphics[scale=.21]{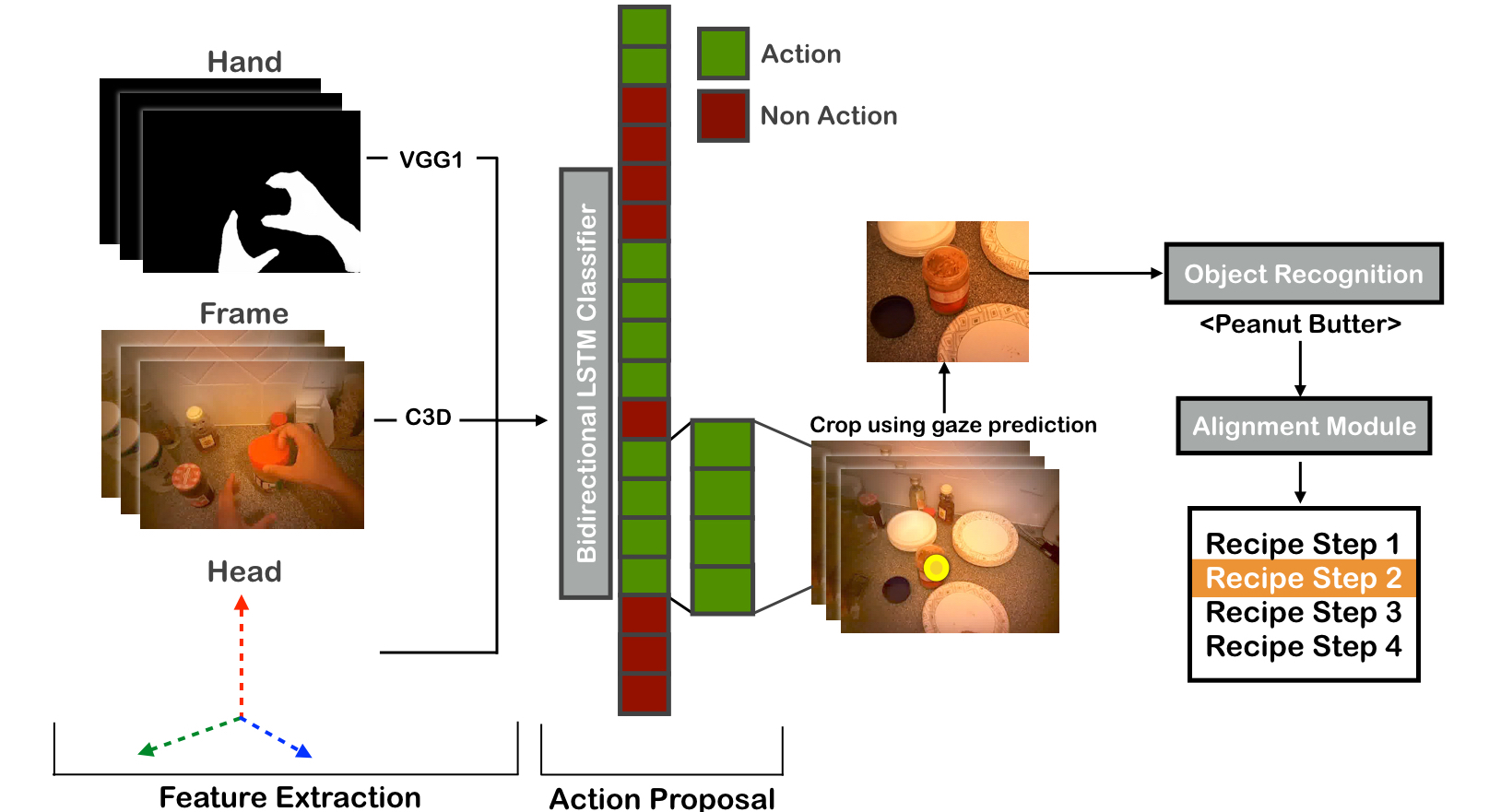}
    \label{full}
  \caption{Video-Recipe Alignment Pipeline: features extracted from head motion, hands and frames, are concatenated and fed into a bidirectional LSTM, which predicts whether a frame contains an action or not. Frames from each action segment are fed into the object recognition network, which builds a object histogram and passes the top k objects to the recipe alignment module. The recipe alignment module parses the recipe and extracts action-object pairs. Then, using the detected objects, action segments are aligned to a recipe step.}
\label{full}
\end{figure*}

\subsection{Generation of Action Proposals}
The first component of our pipeline is a novel method for generating action proposals using egocentric cues. The goal at this stage is to identify which segments of an input video contain actions and which ones do not. We pose this as a supervised frame level classification problem, because the end goal of our system is to align each frame to a single recipe step. We employ a bidirectional LSTM recurrent neural network to perform the classification. We chose this approach because LSTMs store contextual information in their memory cells, allowing the system to store the temporal context of an action, and because bi-directionality propagates information about preceding and succeeding frames. This allows us to exploit the fact that the likelihood that a frame depicts an action is affected by whether the surrounding frames depict actions. The primary novelty in this approach is our success in harnessing egocentric cues in order to segment fine-grained actions which are hard to localize because their short duration. The egocentric features can capture the subtle movement patterns of the hands and head which can indicate the start or end of an action, and which are quite challenging for other methods to capture using more general image feature representations. The results of our experiments provide evidence for this claim.

The instructional videos in for egocentric datasets we use come with frame level annotations which specify if an action is occurring in the frame. If an action is occurring the annotation states which objects are being manipulated and which action class is occuring. The proposal network never uses the action class label, it is trained using only the labels of if an action is occurring. We define a ground truth action segment as a sequence of consecutive frames that have been annotated as having an action occurring. Our bidirectional LSTM network does frame level prediction on whether an action is occurring and is trained using the frame level annotations from the dataset. At test time the network predicts whether or not an action is occurring for each video frame. After making predictions on all frames, we smooth the predictions to eliminate action sequences that are fewer than 10 frames, since action sequences are always longer than this.  
 
The LSTM network is fed both egocentric and spatio-temporal context features. We automatically extract the egocentric features of \emph{head motion} and \emph{hand pose} from the videos in the EGTEA and BEOID datasets, which were captured with head-mounted cameras. Hands are highly discriminative for actions, especially in activities of daily living which require substantial object manipulation. Hands help distinguish action segments, since a subject’s hands are likely to be both visible and in motion when an action is being performed. From each video frame, we segment out the hand from the rest of the image using a deep semantic segmentation method~\cite{long2015fully}. Each segmented hand region is fed into a pre-trained VGG16 network and the last fully connected layer is extracted as the hand feature for the frame. Head motion in egocentric videos corresponds to camera motion. In contrast to the hands, head motion is usually at its peak right before and right after an action. To model camera motion we follow a similar procedure to \cite{li2015delving}. We first match sparse interest points generated by ORB \cite{orb} between adjoining frames. All interest points that lie on the hand mask are deleted. Then we use RANSAC  \cite{ransac} to fit the points to a 9 dimensional homography. We set the homography to zeros for frames with too low of a quality to extract meaningful interest points. The spatio-temporal features we use are the C3D features \cite{tran2015learning} which have been shown to capture both appearance and temporal information. We test the combinations of features in a comprehensive ablation study in \S\ref{sec:action_proposal_eval} and find that the concatenation of the hands, head, and C3D features yield the best performance. 

\subsection{Task-Specific Object Recognition}
\label{sec:obj_reco}
The second module of our system uses the manipulated objects detected in the action segments as the main discriminatory signal for aligning the segment to a recipe step, as described in \S\ref{sec:pipeline}. We found that object names to be discriminative than action names when matching a video segment to a recipe step. The action labels that make up our datasets provide a weaker signal due to the fact that they are fine-grained. For example some frequently occurring actions are ``take'',``put'', and ``close.'' These actions are present in most recipe steps, making them too ambiguous to support accurate alignment. The keys that differentiate the segments are the type of object being manipulated and the order that the objects are manipulated in. Therefore we create a module to identify each object that is manipulated during in the action segments and then use the detections for alignment. 

In order to identify the objects being manipulated in each frame, we train a ResNet-101~\cite{resnet} network on a training split of images extracted from our video corpus. We obtain the training images by extracting and cropping on in every five frames around the point of gaze fixation. We label the image according to the main object annotation for that action segment. The gaze information is obtained from a wearable eye tracker, which is available in both EGTEA and BEOID.

At test time we observe that incorporating egocentric information by cropping the image around the fixation point improved prediction accuracy for most objects. We hypothesize that this is due to subjects almost always looking near the object that they are manipulating. For FP videos that do not contain gaze points, we could potentially use techniques such as \cite{li2013learning} to predict the gaze points. 

We feed every frame inside an action through our network as depicted in Figure \ref{full} to obtain a histogram of object predictions for every action segment. The three top elements of each histogram are passed to the alignment module. This approach gives us robust predictions of the objects that are being manipulated in the action segments, and allows us to filter spurious mistakes and false detections.

\subsection{Recipe Parsing and Alignment}
\label{sec:pipeline}

The final module of our pipeline completes the alignment of each proposed video segment to a recipe step, based on the correspondence between the automatically detected video objects and the text description. Our alignment module is \emph{not trained}, and does not require ground truth manual annotations. As a result, it is scalable and can be applied to diverse datasets. The alignment problem is \emph{challenging} because of the mismatch between the objects that are present in a video and the manner in which objects are referred to in the concise natural language descriptions that comprise a recipe. We address this challenge using tools from natural language processing to identify and extract the key verbs and nouns from the recipe, and to resolve ambiguities and establish references between steps in a recipe.

First, each recipe step is analyzed with the Stanford Dependency Parser \cite{de2006generating}, which identifies the dependency relations between the words in a sentence. The key type of dependency relation that exists between an action and the object of the action, is called a direct object relation. For example, in the sentence ``pour the milk, sugar and coffee into the mug'' the only direct object is from ``pour'' to ``milk.'' From this example, we can see that the direct object relations obtained from the dependency tree will not encompass all of the action-object pairs in the sentence. Therefore, we perform additional processing on the results of the dependency tree. For lists of objects, we extract all of the nouns in the list by identifying the conjunction relations with the direct object noun. In the example sentence, this extracts the nouns ``sugar'' and ``coffee.'' 
In the example, ``milk, sugar and coffee'' are the primary manipulated objects and ``mug'' is a secondary manipulated object. Secondary  objects are noun phrases which have a nominal noun modifier dependency relation with the main action. In the example, the nominal noun modifier relation is between ``pour'', ``in'' and ``mug''. Another challenge arises when pronouns such as ``it'' or ``them'' are used to reference items mentioned in previous steps. When this occurs, we use co-reference resolution to identify the noun that the pronoun is referring to. If an object in a recipe step $x$ is resolved to a previous recipe step $y$, then we create the constraint that the module must align action segments to step $y$ before aligning action segments to step $x$, since the step $y$ is required to complete step $x$. Our parsing algorithm effectively extracts action-object pairs from the recipe and resolves references, as illustrated in Figure \ref{parse}.

The next step is to calculate an \emph{alignment similarity score} between each frame of an action segment and each recipe step, which will be used to produce the final alignment. The score is a weighted average of three distances: the word vector similarity between the primary and secondary manipulated objects and the detected objects, the word vector similarity between the recipe action and the detected objects, and the temporal similarity of the video segment to the recipe step. To obtain word similarity we use the Word2Vec \cite{goldberg2014word2vec} embeddings and calculate the euclidean distance between vectors. Temporal similarity is measured by computing the difference between the ratio of time of the proposed segment to the length of the entire video and the ratio of the number of the recipe step to the total number of recipe steps. This imposes a loose temporal constraint on the alignment. This constraint makes the assumption that while people may do things a few steps out of order, it will be unlikely, for example that the first step preformed will be the last step in the recipe. As a final step, the scores for a segment are weighted using a normal distribution so that frames at the center of the segment contribute more. Note that we discard segments whose similarity score is below a certain threshold, which helps to prune incorrect or irrelevant proposals, since the videos can contain additional actions that do not correspond any recipe steps. We then produce the final alignment by combining the frames in each segment to identify the recipe step with the highest score, and assign the resulting step to all frames in the segment.

The novelty of our alignment module lies in the combination of parsing methods to extract primary and secondary manipulated objects as well as the utilization of word embeddings to infer correlations that are not explicitly stated between the video segments and the recipe steps. For instance the recipe step ``chop the carrots'' implies the actions of getting a cutting board and a knife but does not state them. Through our alignment module's specific use of word vector similarities, we are able to align action segments such as retrieving the cutting board with the correct recipe step. By drawing on the linguistic semantic bank, we can infer what activities are being performed simply by the objects being manipulated and the order they are manipulated in.
\section{Experiments}
\label{sec:experiments}

\noindent\textbf{Datasets.}
We evaluate our system over two egocentric datasets: the Extended GTEA Gaze (EGTEA) dataset  \cite{li2015delving} dataset which is composed of first person cooking videos and the Bristol Egocentric Object Interaction Dataset (BEOID) \cite{beoid} which contains first person videos of scripted daily activities. EGTEA contains 86 unique cooking videos from 32 subjects with an average length of 15 min. There are 7 different recipes, from making a pasta salad to cooking a bacon and eggs breakfast. EGTEA is challenging because the videos are long and complex and the word to text density is quite low (see Figure~\ref{stats}.) While all EGTEA videos were recorded in the same indoor kitchen setting, the manipulated objects differ greatly between recipes, and most recipes require cooking multiple things. Some challenges in the dataset include subjects will cooking multiple foods at the same time and therefore jumping back and forth between recipe steps. BEOID contains 58 videos from 8 participants with an average of one minute. There are 6 different indoor tasks, each performed in a different location, from operating gym equipment to making a cup of coffee. Using the descriptions given to participants, we created an instruction set (recipe) for each task and labeled the ground truth action segments appropriately. We evaluate over BEOID to analyze how our system performs given multiple types of scenes and non-cooking activities. For both BEIOD and EGTEA, we performed additional annotations to support our evaluation by adding the number of the appropriate recipe step to each ground truth action segment. Actions that do not correspond to any recipe step were labelled as such. Note that all performance numbers are obtained using five-fold cross-validation.

\subsection{Action Proposal Evaluation}
\label{sec:action_proposal_eval}
The goal of the action proposal module is to temporally localize actions in videos. In the action proposal module, egocentric features and C3D \cite{tran2015learning} features are fed into a bidirectional LSTM which determines whether or not an action is happening in the frame. Consecutive action frames are classified as a single action segment and then action segments are then smoothed. The single layer bidirectional LSTM is trained for 10 epochs at a learning rate of $1e-4$. The input vector to the network is a 8201 dimensional and we use 300 hidden units. Both EGTEA and BEOID have frame level annotations of actions. The frames in EGTEA videos have a 30:70 split action to non-action and the BEOID has a 40:60 split. 

To measure the accuracy of the proposal system by intersection over union of the predicted and ground truth action segments. We calculate the IOU at three different $\alpha$ thresholds. We compare our action proposal system against the Deep Action Proposal (DAP) system \cite{escorcia2016daps}, which uses C3D features and an LSTM to generate proposals. We retrain the DAP network for both datasets using activations of the last fully connected layer of the C3D network and we extracted the features from our videos at a 8-frame temporal resolution. In our model we use the 4096 dimensional vector extracted from C3D, however for the DAP network we follow their procedure and reduce the features using PCA to 500 dimensions. DAP outputs many proposals along with a confidence score. Our model does not allow for overlapping action proposals, so in order to do a fair comparison we run Non-Maximal Suppression to select the most confident DAP proposals as action segments to test.

Additionally, we perform an extensive ablation study to examine the effectiveness of different types of features which can be seen in Table~\ref{proposal_table}. In the table, it is clear that the addition of the egocentric features significantly increases the IOU score of the proposals. We also tried using traditional features such as improved Dense Trajectories and generic features such as pre-trained VGG network extracted frame features. We observed that improved Dense Trajectories (iDT) \cite{wang2011action} did not work well for our task. We hypothesize this occurred due to over-fitting caused by the large size of the fisher vector encoding of iDTs. Similarly we believe that the C3D + hands + frame feature has a dip in performance because the feature vector is larger and wasn't able to optimize as well given the same amount of training data. 

\begin{table*}[htbp]
\setlength{\tabcolsep}{5pt}
\centering
\caption{Action Proposal Evaluation: IOU values over three different thresholds are given for each proposal method. The results from inputting different feature combinations to the bidirectional LSTM are shown. \textit{frame} denotes pretrained VGG16 features. }

{\small
\begin{tabular}{@{}lccc|ccc@{}}
	\toprule
    Action Proposal Methods  &   \multicolumn{3}{c}{BEOID} &  \multicolumn{3}{c}{EGTEA} \\ \midrule
	 & .5 & .4 & .3 & .5 & .4 & .3 \\ \midrule
	Deep Action Proposals \cite{escorcia2016daps} & 0.1824	& 0.2625 & 0.4129  &  0.1757 & 0.2691 & 0.3966\\
	\textit{Bidirectional LSTM:} &&& \\
	iDT                   & -  & -  & - & 0.1763 & 0.2371 & 0.2952\\
    frame          		  & 0.1846  &  0.2957  & 0.4359 & 0.1647 & 0.3235 & 0.5177\\
    C3D      & 0.3720  &  0.4538  & 0.5712  &	 0.2599 & 0.3501 &  0.4529 \\
	C3D + hands      & 0.5091  &  0.6037  & 0.6832  &  \textbf{0.4051} & 0.5262 & 0.6249 \\
	C3D + hands + frame      & 0.3899  &  0.5114  & 0.6380 & 0.2344 & 0.3750 & 0.6094\\
    C3D + hands + head motion & \textbf{0.5224}   &  \textbf{0.6117} &\textbf{0.7235}  & 
   0.3848 &  \textbf{0.5482}   &\textbf{0.6399}\\ \bottomrule
\end{tabular}
}
\label{proposal_table}
\end{table*}

\subsection{Task-Specific Object Recognition Results}
\label{sec:object_results}
We test our object classification network as a standalone module using five-fold cross-validation on EGTEA Gaze+ and BEOID. We extract one in every five frames from all action segments of the videos and we label each frame as the main task-specific object in the scene. We augment our data by using randomly resized crops spanning aspect ratios of $0.7$ to $1$ and random horizontal flips. At test time, we crop a square of dimensions $400\times400$ around the gaze point and resize it into a square image of size $224\times224$. If gaze points are not obtainable we simply resize the image to $224\times224$.

We use an ImageNet pre-trained ResNet-101 architecture~\cite{resnet} which we fine-tune on each datasets. We train the network until convergence and use a batch size of $64$. The optimization algorithm used was Adam~\cite{kingma2014adam} with learning rate of $3e-4$, weight decay of $1e-4$, $\beta_{1}=0.9$, $\beta_{2}=0.999$ and $\epsilon=1e-8$. We test our fine-tuned object recognition network using five-fold cross-validation on the EGTEA Gaze+ and BEOID datasets and present top-1 and top-3 accuracy. For BEIOD top-1 accuracy was 0.756 and top-3 accuracy was 0.946. For EGTEA top-1 accuracy was 0.542 and top-3 accuracy was 0.732. We observe that task-specific object recognition on the EGTEA dataset is substantially harder than in the BEOID dataset. This is due in part to the long-tail distribution of objects that is present in the EGTEA dataset, with some object classes presenting much larger amounts of instances than others. For example there are $8,361$ images for the ``cucumber'' class whereas ``teabag'' presents $548$ images and ``coffee'' presents only $155$.

\subsection{Video-Recipe Alignment Results}
\label{sec:align_res}

\begin{table}[b]
\caption{Recipe-Video Alignment Accuracy:  AP is given for each alignment method. The last two rows are ablation studies using ground truth (GT) data.}

\label{fullresults}
\setlength{\tabcolsep}{9pt}
\centering
\small
\begin{tabular}{@{}lcc@{}}
    	   & \multicolumn{2}{c}{Average Precision} \\ 	\toprule
       Method   & BEOID & EGTEA  \\ \midrule
	\textbf{(i)} \cite{bojanowski2015weakly} w/ original features & 0.2828    & 0.2169      \\
    \textbf{(ii)} \cite{bojanowski2015weakly} w/ our features & 0.4069    & 0.2292       \\

	\textbf{(iii)} Detected Objects, DAP ~\cite{escorcia2016daps} & 0.2549 & 0.3175    \\
	\textbf{(iv)} Detected Objects, Our Action proposal & \textbf{0.7641}   & \textbf{0.4905} \\ \bottomrule
	\textbf{(v)} Detected Objects, GT action segments        & 0.8272     & 0.5658     \\

	\textbf{(vi)} GT objects, GT action segments           & 0.8425  & 0.6830          \\ \bottomrule
\end{tabular}
\end{table}

\noindent\textbf{Evaluation measure.}
Our system has three modules: action proposal, object recognition and text to action segment alignment. The effectiveness of each module is assessed by substituting it with ground truth results Table~\ref{fullresults}.
To evaluate the final alignment, we use the same metric as~\cite{bojanowski2015weakly}, which computes the precision of the frame level alignment predictions. The reported score is the averaged score of all the videos in the test set. 

\noindent\textbf{Baseline.}
We compare our alignment module to the recent alignment method of~\cite{bojanowski2015weakly}, which is also designed to align video and text. To do a fair comparison of alignment modules we test the baseline both with our deep features and the original hand-crafted features employed in the paper~\cite{bojanowski2015weakly}, which are bag-of-words representations of improved dense trajectories~\cite{wang2011action} for HOG, HOF and MBH channels. To represent the recipe steps, we use simple bag-of-words of ROOT verbs as it was the best performing strategy (as observed in~\cite{bojanowski2015weakly}). Hyper-parameters are chosen by cross validation over our splits.

\noindent\textbf{Methods.}
All results are reported in Table~\ref{fullresults}.
\textbf{(i)} corresponds to the baseline~\cite{bojanowski2015weakly}, which \emph{does} not use action proposal techniques.
\textbf{(ii)} corresponds to our pipeline but uses the action proposal method introduced in~\cite{escorcia2016daps}.
\textbf{(iii)} corresponds to our full pipeline.
Finally the in ablation studies, \textbf{(iv)} and \textbf{(v)}, we replace parts of our pipeline with the ground truth information in order to assess where improvements have to be made.
More precisely \textbf{(iv)} uses ground truth action segments and \textbf{(v)} uses both ground truth action segments and objects.

\noindent\textbf{Result analysis.}
We can view the results that use ground truth (GT) labels as upper bounds as well as an ablation study of the individual modules of our system. The result of using the GT object labels and the GT action segments \textbf{(vi)} tests only the alignment module. We can see that there is an 15.25\% difference in AP between the datasets. Despite the BEOID being taken in multiple different settings it still outperforms EGTEA for alignment. We can attribute this to the higher density of the recipes, directly proving the difficulty of the task of sparse text to video alignment.

The poor performance of the alignment baseline shown in \textbf{(i)} and \textbf{(ii)} can be explained by the fact that the task considered here is more challenging than usual text-to-video alignment tasks due to our datasets' high text sparsity. More priors have to be provided in order to achieve  better performance for the baseline. This is demonstrated with \textbf{(iv)} which uses action proposal in order to reduce the search space for actions. We can see a clear difference in performance between methods \textbf{(i)} and \textbf{(ii)}, this difference corroborates the power of the egocentric cues in our features. When comparing the action proposal baseline \textbf{(iii)} and our system \textbf{(iv)}, we can see a extremely large boost in performance ($+35.34\%$) induced by our new action proposal which is more suited for FP videos. Unlike the other methods, baseline \textbf{(iii)} performs worse on BEOID than EGTEA. When looking at the results on individual videos in BEOID  we found the bad performance was due to failure on the short videos. The method was producing too many proposed segments for the short videos leading to a low AP score that was not reflected in the low IOU.

We can gain some interesting insights by looking at the experiments using ground truth labels. Using ground truth action segments \textbf{(iv)}, brings a relatively small improvement (BEOID: $6\%$, EGTEA: $7\%$) over our generated action proposals \textbf{(iii)}, thus showing that our action proposal system works extremely well for this task. The difference between using ground truth objects \textbf{(v)} and using object recognition \textbf{(iv)} is also small, meaning that our alignment system is able to handle mislabeled objects without breaking down. The the drop between \textbf{(v)} and \textbf{(iv)} is smaller for BEIOD than for EGTEA because our object recognition model achieves better performance on the BEOID dataset described in \S\ref{sec:object_results}.
\vspace{-.2cm}

The performance of our upper bound for the alignment \textbf{(v)}, demonstrates the difficulty of the video-to-recipe alignment task. The challenge arises from the sparse nature of the recipe instructions. A recipe step gives an overall description of some activity to complete, however it does not explicitly state the individual actions and objects required to complete this goal. For instance a recipe step such as ``scramble 2 eggs over medium heat'' must be mapped to the numerous actions of cracking two eggs, mixing them and putting them on a pan on the stove. We aim to learn these mappings and we supplement the learning with the semantic knowledge of language. The use of word embeddings in our alignment pipeline allows the system to extract contextual knowledge and find conceptual similarities between the recipe step and detected objects.

\noindent\textbf{Evaluation of Alignment Similarity Score}.
The final module of our pipeline completes the alignment by assigning video segments to the recipe steps based on a similarity score. We do an ablation on the similarity measurement in Table~\ref{aligmentstudy} to show how different factors affect the measurement. 
\vspace{-.2cm}
\begin{table}[htbp]
\caption{Similarity score ablation study for the alignment module on our system: method \textbf{iv}.}
\label{aligmentstudy}
\setlength{\tabcolsep}{9pt}
\centering
\small
\begin{tabular}{@{}lcc@{}} & \multicolumn{2}{c}{Average Precision} \\ 	\toprule
     Similarity Score Type   & BEOID & EGTEA  \\ \midrule
	 Full Similarity Score & 0.7641   & 0.4905 \\ \hline
	 Only Temporal Similarity & 0.4272   & 0.2596 \\ \hline
	 Only Word Semantic Similarity & 0.3981  & 0.2310 \\ \hline

\end{tabular}
\end{table}
\vspace{-.2cm}

\begin{figure*}[t]
  \centering
    \includegraphics[scale=.25]{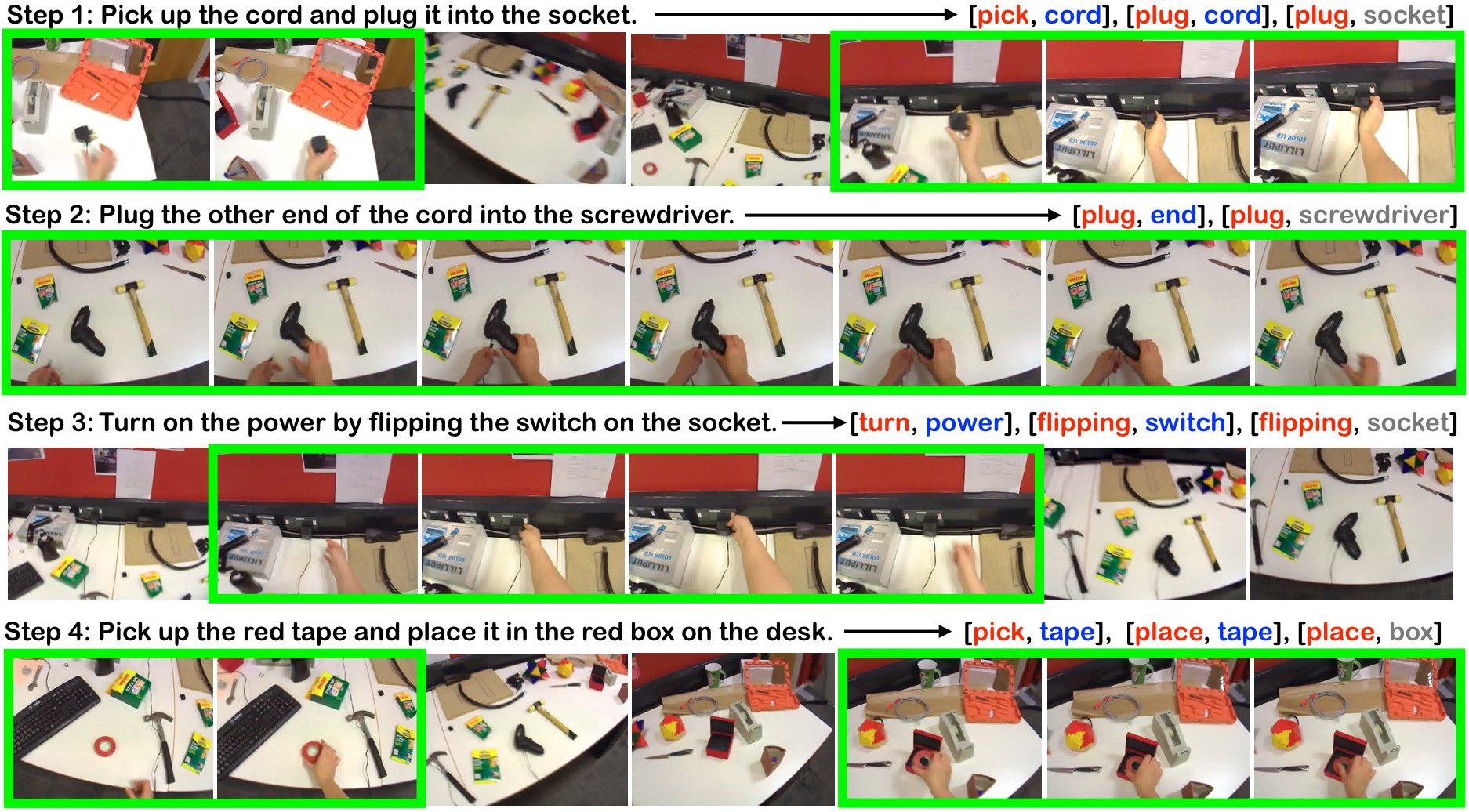}
    \label{end}
  \caption{Alignment results for BEOID. Each recipe step is linked to extracted action-object pairs, with actions in red, primary objects in blue, and secondary objects in gray. Green boxes identify video segments which have been aligned to the step.}
  \label{parse}
\end{figure*}

\section{Conclusions}
\label{sec:conclusion}
In this work we propose a novel a challenging task of recipe to video alignment. We then present a comprehensive method to align fine-grained actions in first person video to sparse recipe steps, shown in Figure~\ref{parse}. Our method is a three-step pipeline which first predicts the onset and offset of action segments in the video, then retrieves task-specific objects that are manipulated during the action and finally uses this information to align the segment to a recipe step using natural language processing techniques. We present results of our recipe alignment pipeline on the new EGTEA dataset and the Bristol Egocentric Object Interaction dataset. For both datasets the system achieves an improved accuracy compared to a recent video and text alignment technique. Additionally, we have shown that by leveraging egocentric cues such as hand masks and head motion we can substantially improve upon state-of-the-art action proposal methods in first person videos.
\\
\\
\\
\\
\\
\\
\\

\end{document}